\documentclass[sigconf]{acmart}

\let\mathscr\relax
\usepackage{algorithm}
\usepackage[noend]{algpseudocode}
\usepackage{amsmath}

\AtBeginDocument{%
  }

\setcopyright{acmcopyright}

\acmConference[ICONS]{International Conference on Neuromorphic Systems (Poster)}{2023}{Santa Fe, New Mexico, USA}

\begin{document}

\title{General-purpose Dataflow Model with Neuromorphic Primitives}

\author{Weihao Zhang}
\email{zwh18@mails.tsinghua.edu.cn}
\orcid{0000-0002-9301-8538}
\affiliation{%
  \institution{Center for Brain-Inspired Computing Research (CBICR), Tsinghua University}
  \city{Beijing}
  \country{China}
}

\author{Yu Du}
\email{duyu20@mails.tsinghua.edu.cn}
\orcid{0000-0002-9301-8538}
\affiliation{%
  \institution{Center for Brain-Inspired Computing Research (CBICR), Tsinghua University}
  \city{Beijing}
  \country{China}
}

\author{Hongyi Li}
\email{hy-li21@mails.tsinghua.edu.cn}
\orcid{0000-0002-7495-9930}
\affiliation{%
  \institution{Center for Brain-Inspired Computing Research (CBICR), Tsinghua University}
  \city{Beijing}
  \country{China}
}

\author{Songchen Ma}
\email{msc19@mails.tsinghua.edu.cn}
\affiliation{%
  \institution{Center for Brain-Inspired Computing Research (CBICR), Tsinghua University}
  \city{Beijing}
  \country{China}
}

\author{Rong Zhao}
\authornote{Corresponding author.}
\email{r_zhao@tsinghua.edu.cn}
\affiliation{%
  \institution{Center for Brain-Inspired Computing Research (CBICR), Tsinghua University}
  \city{Beijing}
  \country{China}
}

\renewcommand{\shortauthors}{Weihao Zhang, Yu Du, Hongyi Li, Songchen Ma, Rong Zhao.}

\begin{abstract}
    Neuromorphic computing exhibits great potential to provide high-performance benefits in various applications beyond neural networks. However, a general-purpose program execution model that aligns with the features of neuromorphic computing is required to bridge the gap between program versatility and neuromorphic hardware efficiency. The dataflow model offers a potential solution, but it faces high graph complexity and incompatibility with neuromorphic hardware when dealing with control flow programs, which decreases the programmability and performance. Here, we present a dataflow model tailored for neuromorphic hardware, called neuromorphic dataflow, which provides a compact, concise, and neuromorphic-compatible program representation for control logic. The neuromorphic dataflow introduces "when" and "where" primitives, which restructure the view of control. The neuromorphic dataflow embeds these primitives in the dataflow schema with the plasticity inherited from the spiking algorithms. Our method enables the deployment of general-purpose programs on neuromorphic hardware with both programmability and plasticity, while fully utilizing the hardware's potential.
\end{abstract}




\maketitle

\section{Introduction}
In recent years, Neuromorphic computing (NC) has attracted much attention as an alternative computing to von Neumann architecture in computing\cite{roy2019towards}. Inspired by the biological brain, NC demonstrates the advantages of ultra-low latency and high energy efficiency in both neuroscience-oriented and intelligent-oriented applications due to its in-situ computing, event-driven processing patterns, and direct utilization of the physical circuit’s functionalities. NC chips typically adopt a many-core architecture that incorporates digital or analog crossbars co-located storage for massively parallel processing. Network-on-chip with routers is used to connect these processing units\cite{li2023brain}. Till now, NC has mostly been used for domain-specific applications. However, there is now an aspiration to extend its high efficiency to more versatile applications.

Efforts from various perspectives are underway to achieve this goal. Theoretically, the NC has been proven to be Turing Complete with corresponding computational model\cite{date2022neuromorphic}. In practice, some works have used NC infrastructures to support non-neural network applications\cite{aimone2022review}, such as solving partial differential equations through random walking\cite{smith2022neuromorphic} and addressing traditional NP-Hard problems\cite{davies2021advancing}. Some neuromorphic chips have also explored the mutual scheduling mechanism between neuromorphic execution activities, replacing central processing units (CPU) to enhance flexibility\cite{ma2022neuromorphic}, replacing the scheduling responsibility of the CPU. Additionally, programming frameworks and compilers have been developed to provide higher-level abstractions for hardware-agnostic programming with portability\cite{aimone2019composing, LAVA}. 


To support the increasing demand for applications and advancements in theories, a neuromorphic program execution model that can unify the representation of general programs with hardware execution abstraction is highly needed. Several mature models have been proposed for domain-specific NC, such as Corelet for TrueNorth\cite{amir2013cognitive} and Rivulet for Tianjic\cite{ma2022neuromorphic}. In terms of general-purpose models, there are efforts to utilize the general expressivity or approximation capability of spiking neurons to either construct or approximate general programs, such as neural engineering framework\cite{eliasmith2003neural}, Fugu\cite{aimone2019composing}, and neuromorphic completeness representation\cite{zhang2020system}. However, these approaches require an extensive number of neurons to construct even a single control operation, and the high-precision low-redundancy approximation for general control is still challenging to achieve general approximation. On the other hand, the dataflow model\cite{dennis1974data}, which is Turing complete and shares similarities with high-level features of NC, has been explored for brain-inspired representation\cite{qu2017parallel}. However, the existing dataflow model has a complex representation for control logic and is incompatible with most NC chips. Moreover, the conventional dataflow has less plasticity than neuromorphic models, limiting its learning ability. To this extent, the main contributions of this paper are:

\begin{enumerate}
\item[1)] We analyze the control logic of von-Neumann programs from a neuromorphic perspective and devise the "where" and "when" primitives.
\item[2)] We propose a concise, neuromorphic compatible, programmable, and learnable neuromorphic dataflow model for general programs with control flows. 
\end{enumerate}


\section{Conventional Dataflow Model}

\begin{figure}[t!]
  \centering
  \includegraphics[scale=0.34, keepaspectratio]{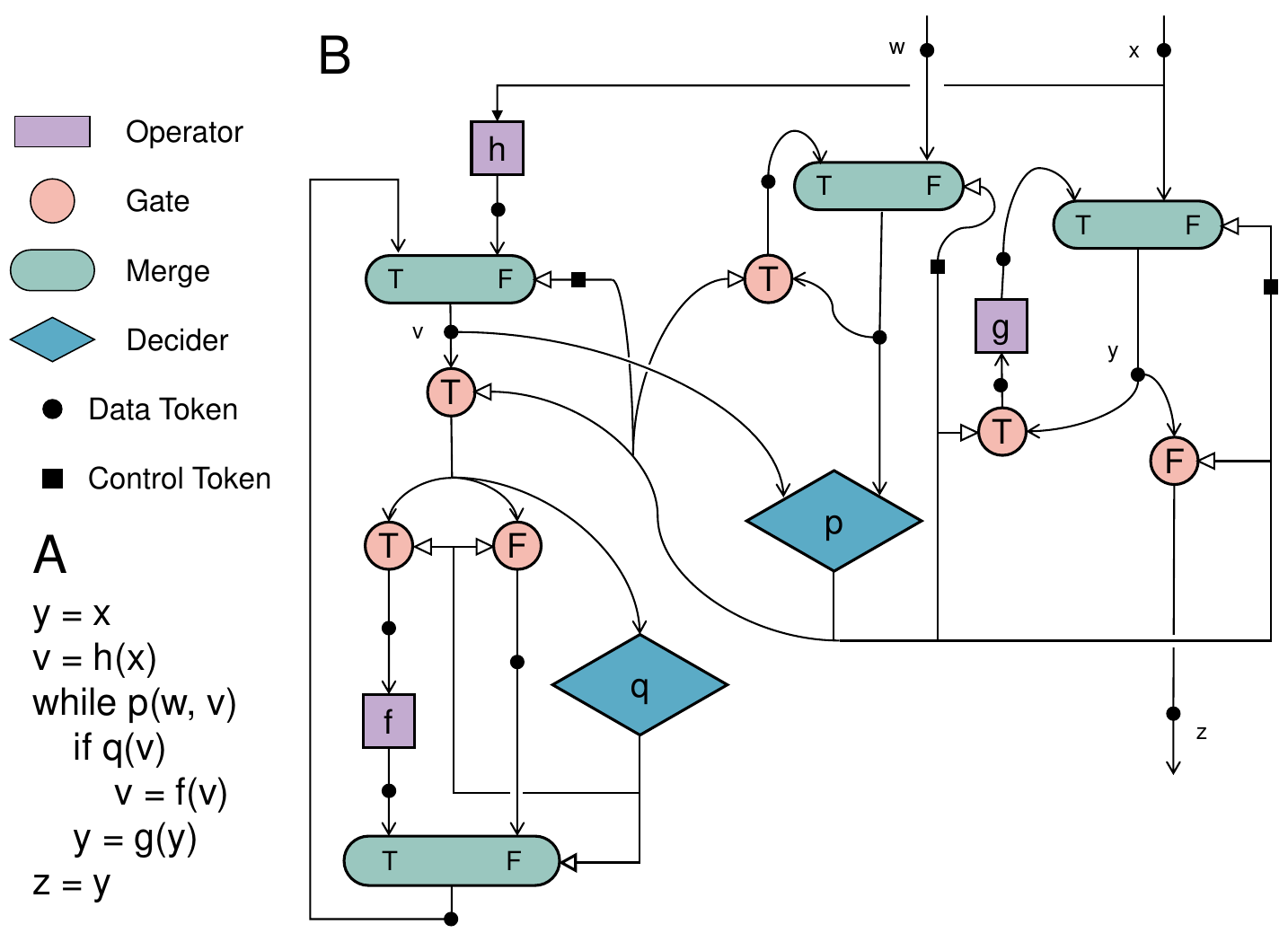}
  \caption{A basic demonstration of conventional dataflow. (A) A program segment written in Algol-like syntax designed for von-Neumann architecture. (B) The equivalent dataflow representation of program A. This example is taken from \cite{dennis1974data}}
  \label{fig:conventional_dataflow}
\end{figure}

A dataflow model is a directed graph in which vertices are called actors and edges are called arcs. The execution of an actor’s operation is initiated by an event known as a token. Arcs transfer tokens between actors, where input tokens are consumed by an actor to generate output tokens based on firing rules. Typically, the dataflow model contains data tokens that denote arbitrary values, and control tokens that represent true/false values. A basic example of the dataflow model is presented in Fig. \ref{fig:conventional_dataflow}, which include operators representing integrated functions with one or multiple input data arcs, and fire output data tokens when all input tokens are available. Deciders, representing predicate logic, require one or multiple data tokens as input and fire a control token when all inputs are available. The true/false gates permit data tokens to pass through when receiving a true/false control token, while the merges let the data token pass through the true side when receiving a true control token, and vice versa.

In the conventional dataflow model, the presence of gates, merges, and arcs that carry control tokens significantly increases the complexity of the graph with control logic. In Fig. \ref{fig:conventional_dataflow}, for example, there are 10 gates or merges for just one "while" logic and one "if" logic. These fine-grained and irregular operators in the dataflow model render it unsuitable for neuromorphic hardware and may cause a mismatch between the dataflow parallelism and fixed hardware parallelism.

\section{Neuromorphic Dataflow Model}

\begin{figure}[t!]
  \centering
  \includegraphics[scale=0.35, keepaspectratio]{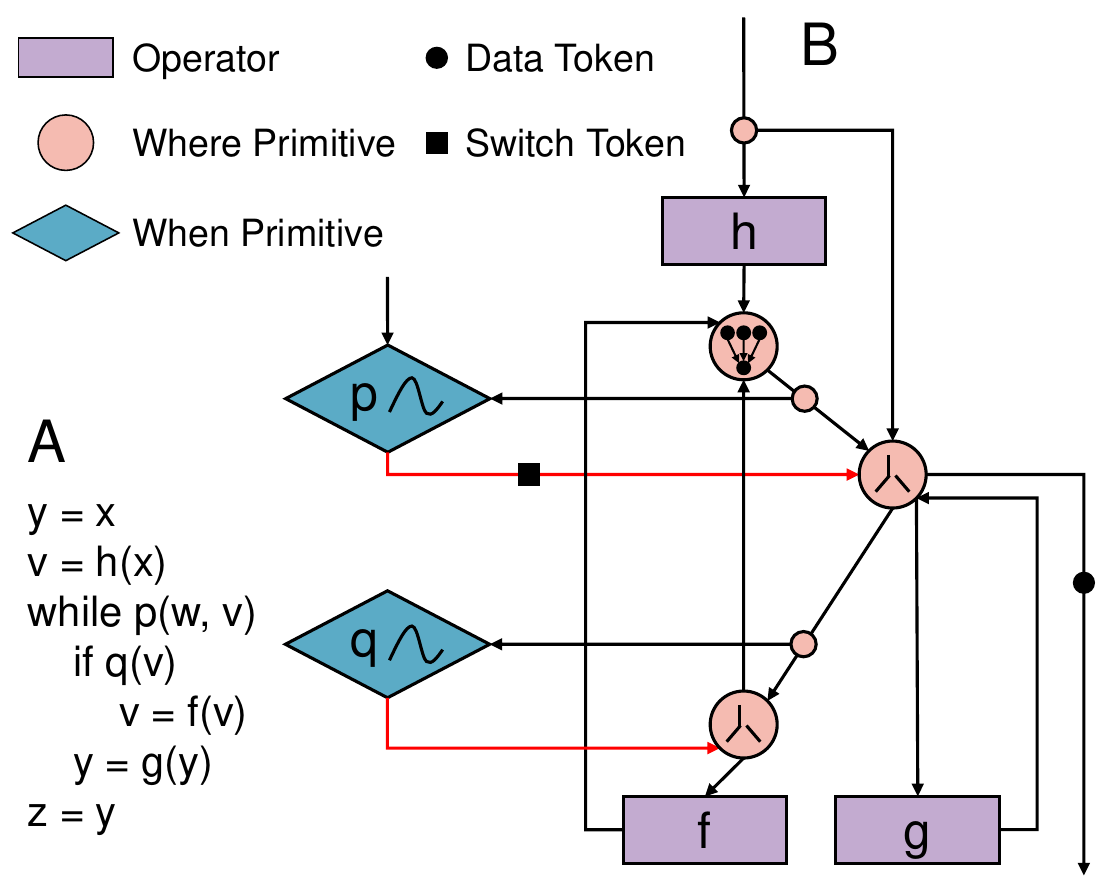}
  \caption{A basic demonstration of neuromorphic dataflow. (A) Same program in Fig. \ref{fig:conventional_dataflow} (A). (B) The equivalent neuromorphic dataflow model of program A.}
  \label{fig:neuromorphic_dataflow}
\end{figure}

To address the above issues, we design a neuromorphic dataflow model (NDF). Classical programs utilize control logic such as conditions, loops, or gotos to manipulate programs based on the program counter (PC). In contrast, the dataflow model utilizes tokens to trigger operations via an event-driven mechanism. Control is realized in a predicate-decision decoupled framework with control tokens. The objective of NDF is to further abstract the concept of "control" from a neuromorphic perspective. Drawing on the dataflow model and neuroscience, such as the gating mechanism in neuronal circuits\cite{luo2021architectures}, the control logic in NDF involves two aspects: \textbf{where} tokens are directed and \textbf{when} tokens are generated. Following this philosophy, we designed \textit{where and when primitives} in NDF.

\begin{figure}[h]
  \centering
  \includegraphics[scale=0.35, keepaspectratio]{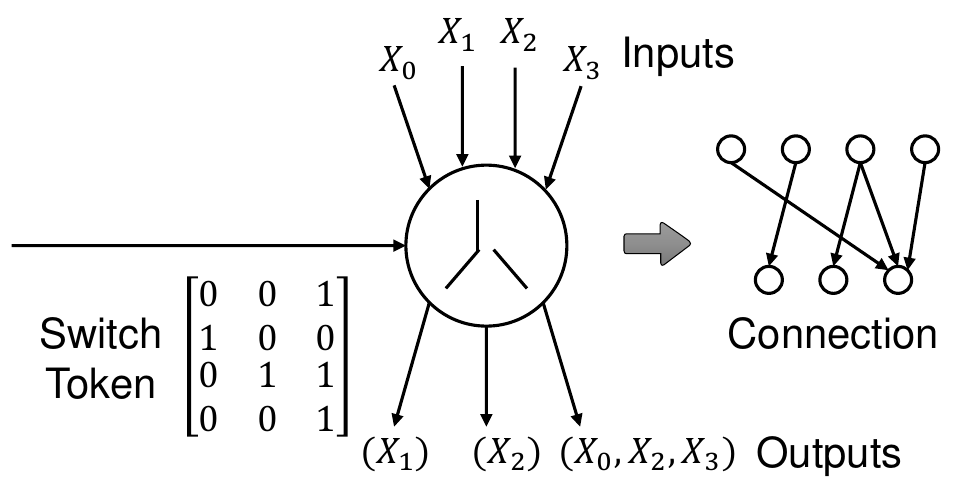}
  \caption{A \textit{dynamic where primitive} takes switch token as input to change its connections of data in/out arcs.}
  \label{fig:where_primitive}
\end{figure}

\subsection{Dataflow Model with Where Primitive}

The \textit{where primitive} replaces gates and merges to create a concise dataflow. It can be viewed as a multi-switch as shown in Fig. \ref{fig:where_primitive}. The \textit{where primitive} has $m_{where}, m_{where}>=1$ data token inputs and $n_{where}, n_{where}>=1$ data token outputs, and redirects input data tokens to output with a specific connection pattern. A \textit{static where primitive} has fixed connectivity, whereas a \textit{dynamic where primitive} relies on another input switch token. The switch token is a specially designed token that determines the connections between inputs and outputs and is equivalent to a $n_{where} \times m_{where}$ adjacency matrix. Each firing of the \textit{dynamic where primitive} will consume a switch token and transfer input data tokens to corresponding output arcs.

With \textit{where primitives}, the complexity of the original dataflow is greatly reduced. Shown in Fig. \ref{fig:neuromorphic_dataflow}, the NDF for the same program in Fig. \ref{fig:conventional_dataflow} only has 8 actors (ignoring the two-to-one data link), including one \textit{static where primitive} with a constant three-to-one connection and two \textit{dynamic where primitives} that controlled by the \textit{when primitives}.

\subsection{Dataflow Model with When Primitive}

\begin{figure}[h]
  \centering
  \includegraphics[scale=0.46, keepaspectratio]{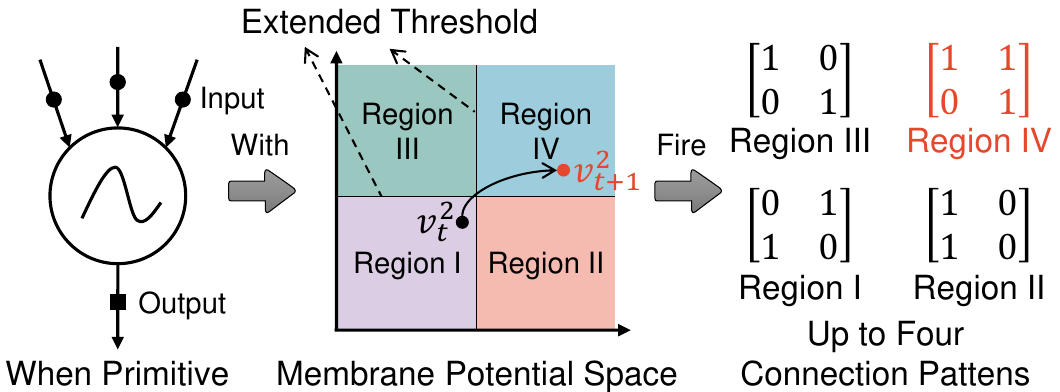}
  \caption{The firing rule of \textit{when primitive} with two-dimensional membrane potential and four separate regions.}
  \label{fig:when_primitive}
\end{figure}

The \textit{when primitive} generates switch tokens to determine the connectivity of the \textit{dynamic where primitive}. A spiking neuron with temporal richness is adopted to replace the original decider. For instance, the predicate $3x - 2 < y$, i.e. $3x - y < 2$ can be modeled as a leaky integrate-and-fire (LIF) neuron that takes two inputs $x$ and $y$ whose weights are $3$ and $-1$ respectively, and a threshold of $2$. If the statement is true, the spiking neuron fires a spike that serves as a switch token for representing the connection pattern (which can be stored in advance) under the true situation.

\begin{figure}[b]
  \centering
  \includegraphics[scale=0.75, keepaspectratio]{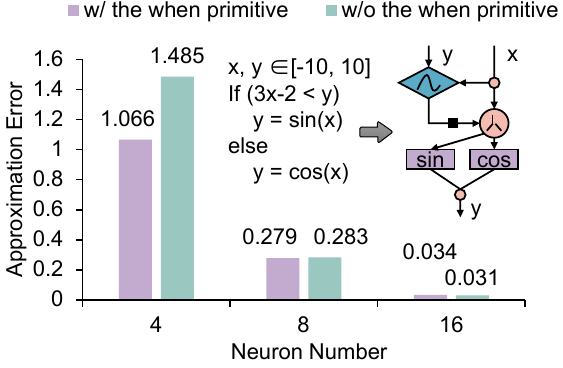}
  \caption{Approximation error of a simple program.}
  \label{fig:error}
\end{figure}

\begin{figure*}[hbt]
  \centering
  \includegraphics[scale=0.45, keepaspectratio]{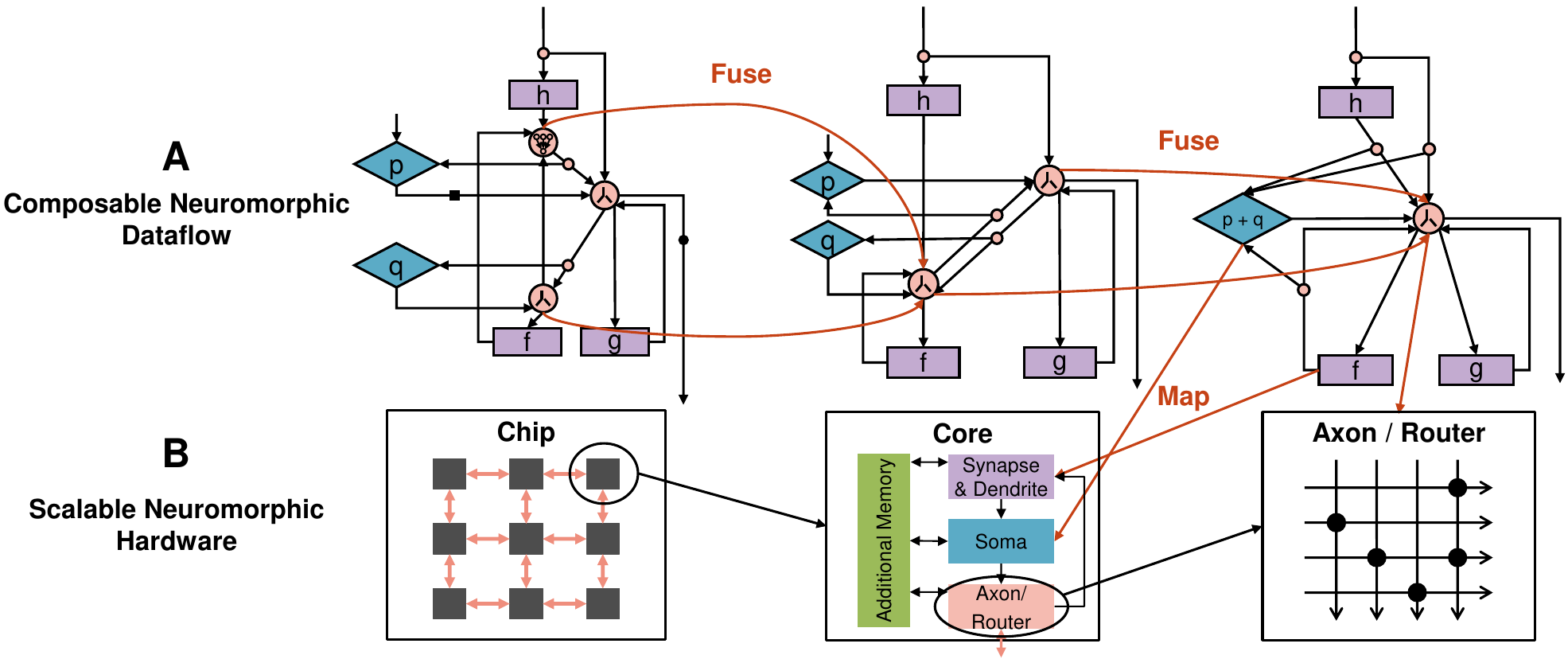}
  \caption{The compatibility between neuromorphic dataflow and neuromorphic hardware. (A) Composable NDF with different granularity. (B) One of the representative neuromorphic hardware architectures with scalable hierarchy\cite{zhang2020system}.}.
  \label{fig:fuse_map}
\end{figure*}

The NDF may have instances where the \textit{where primitive} has multiple possible connection patterns. In such cases, the corresponding \textit{when primitive} should have an output type beyond just spike and not spike. Thus, we introduce a modified spiking neuron for \textit{when primitives}. The modified spiking neuron has a $k$-dimensional membrane potential $v^k$, inputs vector $I$ with length $n_{when}$, and a weight matrix $W$ with size $n_{when} \times k$. The updated rule for membrane potential is:

\begin{equation}
v^k_{t+1} = f \left( v^k_t + I^TW \right)
\end{equation}

Here, we introduce a function $f$ to increase the non-linearity and extend the traditional threshold concept to the separation of the multi-dimensional membrane potential space. Specifically, when the value of $v^k_t$ is within a particular region, the neuron fires a corresponding token for a particular connection pattern. Accordingly, the \textit{when primitive} with $m_{when}$ separate regions can accommodate up to $m_{when}$ connection patterns of the \textit{where primitive}. Fig. \ref{fig:when_primitive} illustrates such a spiking neuron with a 2-dimensional membrane potential and four separate regions within the membrane potential space. The membrane potential shifts among these regions depending on the input. This type of neuron can be regarded as a multi-dimensional state machine in continuous space with the membrane potential updated rule as the state transition equation. 

\section{Hardware Compatibility of Neuromorphic Dataflow Model}

\textbf{The composability of \textit{where primitives}.} The multi-switch structure of \textit{where primitives} makes them compatible with the 2D-mesh router implementations that are widely adopted by neuromorphic hardware\cite{ji2019fpsa}. Alternatively, it can be mapped on a network-on-chip system with fine-grained functional bio-plausible routing protocols\cite{ma2022neuromorphic, furber2014spinnaker}, allowing for the fusion of multiple adjacent \textit{where primitives} into one larger \textit{where primitive}. This enables the adjustment of the granularity of the \textit{where primitives} to achieve improved load-balance on many-core neuromorphic chips, as shown in Fig. \ref{fig:fuse_map}. 

\textbf{The plasticity of \textit{when primitives}.} \textit{When primitives} can be mapped on the soma module (or a modified soma module from co-designing), which is primarily responsible for non-linear functions or dynamic procedures\cite{indiveri2011neuromorphic}. The \textit{when primitive} has a learning ability, through STDP\cite{song2000competitive} or BP with surrogate gradient functions\cite{neftci2019surrogate}, to approximate the target functionality. By training the NDF as a whole with other neuromorphic operators, through the \textit{when primitive}, the overall precision can be improved. We demonstrate this 
through an experiment. 

The $sin$ and $cos$ in the target program in Fig. \ref{fig:error} are each approximated by an MLP with 4, 8, and 16 neurons of the hidden layer. Without the when primitive, the approximation of $sin$ and $cos$ are trained independently. Conversely, the NDF is trained as a whole with the surrogate gradient of the when primitive. As shown in Fig. \ref{fig:error}, using the \textit{when} primitive can reduce the approximation error of NDF when the number of neurons is small (4 and 8), but little space is left to reduce the overall error. However, when the number of neurons is enough to approximate $sin$ and $cos$ functions precisely.

\section{Conclusion}

We present a program execution model that combines the dataflow schema with neuromorphic-compatible primitives, allowing for the practical programming and deployment of a wide range of applications on Turing-complete neuromorphic hardware. To achieve this, we design the \textit{where primitive} to direct tokens and \textit{when primitive} to control when tokens fire. By incorporating these primitives, our NDF model achieves a compact, concise, and interpretable control-logic representation. The NDF also exhibits compatibility with neuromorphic hardware that has plasticity and multi-grained composability. Our work introduces a dataflow perspective to practical general-purpose neuromorphic computing, providing the potential to combine brain-level efficiency and CPU-level versatility.

\begin{acks}
This work was partly supported by National Nature Science Foundation of China (nos. 61836004 and 62088102).
\end{acks}

\bibliographystyle{ACM-Reference-Format}
\bibliography{NDF_ref}

\end{document}